\documentclass[a4paper, 10pt, conference]{ieeeconf}      % Use this line for a4 paper
\IEEEoverridecommandlockouts                              % This command is only needed if 
\usepackage{hyperref}
\usepackage{orcidlink}
\usepackage{graphicx}
\usepackage{array}
\usepackage{setspace}
\usepackage{caption}
\usepackage{etoolbox}
\makeatletter
\patchcmd{\@makecaption}
  {\scshape}
  {}
  {}
  {}
\makeatletter
\patchcmd{\@makecaption}
  {\\}
  {.\ }
  {}
  {}
\makeatother

\title{\LARGE \bf Plant-inspired behavior-based controller to enable reaching\\in redundant continuum robot arms}

\author{Enrico Donato$^{1}$ \orcidlink{0000-0002-8844-5279}, Yasmin Tauqeer Ansari$^{1}$ \orcidlink{0000-0002-5967-5067}, Cecilia Laschi$^{2}$ \orcidlink{0000-0001-5248-1043} and Egidio Falotico$^{1}$ \orcidlink{0000-0001-8060-8080}% <-this % stops a space
\thanks{This work was supported by the European Union’s H2020 Research and Innovation Programme through GROWBOT Project under Grant 824074.}% <-this % stops a space
\thanks{$^{1}$E. Donato, Y.T. Ansari and E. Falotico are with The BioRobotics Institute, Sant'Anna School of Advanced Studies, 56025 Pontedera (PI), Italy and with the Departement of Excellence in Robotics \& AI, Sant'Anna School of Advanced Studies, 56125 Pisa, Italy {\tt\small \{e.donato, y.ansari, e.falotico\}@santannapisa.it}}%
\thanks{$^{2}$C. Laschi is with the Department of Mechanical Engineering, National University of Singapore, 127575 Singapore, on leave from the BioRobotics Institute, Sant'Anna School of Advanced Studies, 56125 Pisa, Italy. {\tt\small mpeclc@nus.edu.sg}}%
}

\begin{document}
    \maketitle
    %\thispagestyle{empty}
    %\pagestyle{empty}
    
    %%%%%%%%%%%%%%%%%%%%%%%%%%%%%%%%%%%%%%%%%%%%%%%%%%%%%%%%%%%%%%%%%%%%%%%%%%%%%%%%
    \begin{abstract}
        Enabling reaching capabilities in highly redundant continuum robot arms is an active area of research.  Existing solutions comprise of task-space controllers, whose proper functioning is still limited to laboratory environments. In contrast, this work proposes a novel plant-inspired behaviour-based controller that exploits information obtained from proximity sensing embedded near the end-effector to move towards a desired spatial target. The controller is tested on a 9-DoF modular cable-driven continuum arm for reaching multiple set-points in space. The results are promising for the deployability of these systems into unstructured environments.
    \end{abstract}  

    %%%%%%%%%%%%%%%%%%%%%%%%%%%%%%%%%%%%%%%%%%%%%%%%%%%%%%%%%%%%%%%%%%%%%%%%%%%%%%%%
    \section{Introduction}
        Modular continuum robot arms refer to a new generation of bio-inspired manipulators comprised of a redundant arrangement of actuation units, each constituted of a combination of lightweight flexible actuators \cite{c1}. The resulting systems can continuously deform at any point along its length, thereby, enabling compliance towards externally applied loads. These desirable properties are particularly effective in manipulating natural environments, with a high-degree of human-robot safety, which are functionalities beyond the scope of their rigid counterparts \cite{c2}. 
        
        A commonly identifiable task for these systems is to reach a desired spatial location. Significant advancements have been made in this regard through the development of kinematic task-space controllers to determine the appropriate input activations to achieve the desired task \cite{c2}. In general, the functioning of such controllers relies on computational formulations that can create a valid mapping between task-space and actuator-space, and have been broadly achieved in three ways: (i) analytical approximation; (ii) machine learning, and; (iii) online estimation algorithms. However, the proper functioning of these controllers generally relies on vision-feedback which limits their validity within laboratory environments, restricting the deployability of these systems in natural and dynamic environments. This article is the first attempt to overcome this unaddressed limitation, and extend the reach of these systems to unstructured environments.
        
        This objective can be achieved by taking cue from plants which, despite their seemingly sessile nature, survive in almost all habitats by actively negotiating with a wide range of terrain using movement strategies, based on growth \cite{c3}. Interestingly, due to the lack of a central nervous system, these movement capabilities arise because of sophisticated forms of decentralized computing mechanisms \cite{c4} that exploit information from embedded biosensors. This motivates the investigation of a novel paradigm of control strategies for continuum robot arms that do not rely on establishing mappings between robot operating spaces. In the spectrum of artificial intelligence, decentralized control is implemented through distributed architectures, which comprise of autonomous computing agents that can act independently and asynchronously, thereby, offering a computationally efficient approach to problem solving \cite{c5,c6}. In particular, distributed architectures were pioneered through the development of the behaviour-based controllers \cite{c7}, which comprises of task-achieving modules, known as behaviours, that achieve a specific objective in response to sensory inputs and/or internal state. Then, the overall functionality is achieved through the collective interaction of individual behaviors with each other, as well as the environment. 
        
        This work takes inspiration from the ability of rod-like organs (shoots and roots) of climbing plants to reach nutrients by navigating through competitive environments. In particular, this ability arises from growth-driven movements \cite{c8}: (i) nastic - due to internal drivers, such as the inherent periodic movement called circumnutations which is associated with search processes. and; (ii) tropic – is the response of the plant in the direction of a stimulus, such as a plant shoot growing toward a source of light or away from the direction of gravity, etc. This lays the foundation to develop a novel behaviour-based controller that can generate reaching capabilities in redundant continuum arms through a bottom-up arrangement of behavioural modules emulates growth-driven movements, and is the main contribution of this work. 
        
        Section \ref{robotic_platform} presents the robotic platform followed by the formulation of the control framework in Section \ref{control_framework}. The performance of the controller is validated on a 9-DoF modular cable-driven continuum arm for reaching multiple set-points in space in Section \ref{results}. The same controller can also enable soft arms to perform reaching tasks in unstructured environments, provided they have the same structural characteristics and movement constraints, improving the preliminary work discussed in \cite{c9}. Finally, Sections \ref{discussion} and \ref{conclusion} comprise of a discussion of the obtained results with future outlooks.

    \section{Robotic Platform} \label{robotic_platform}
\begin{figure}[h!]
            \centering
            \includegraphics[width=1\columnwidth]{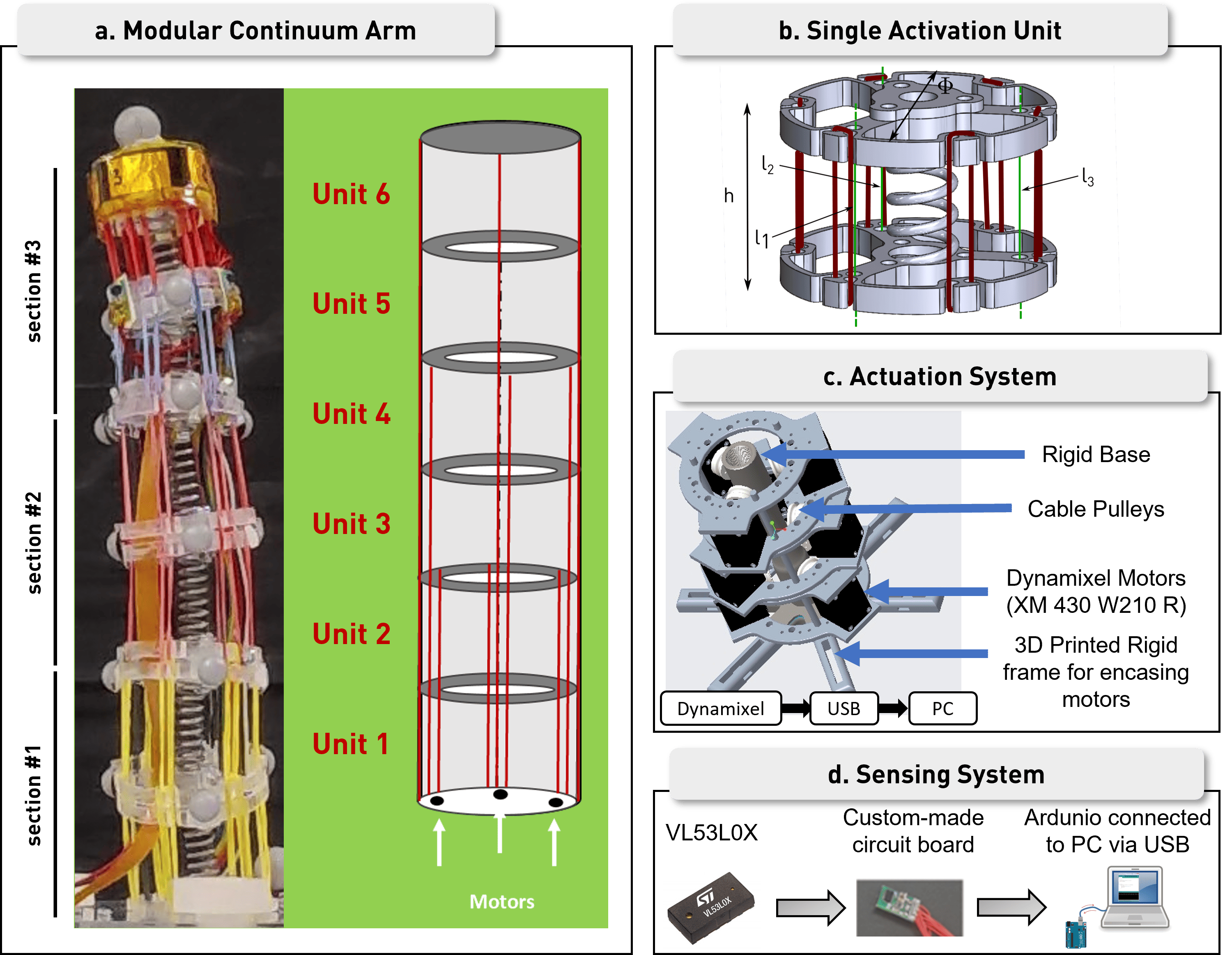}
            \caption{The (A) modular continuum robot arm comprises of a concatenation of (B) activation units. The overall system is a combination of a pair of activation units that are independently activated, and is mounted on a (C) custom-designed rigid base. The system monitors the environment through a distributed arrangement of (D) proximity sensors placed in distal portion of the arm.}
            \label{fig:robotic_platform}
        \end{figure}

        This work employs a lightweight modular continuum arm. A single activation unit, as shown in Fig.\ref{fig:robotic_platform}(B), is delimited by two acrylic plates that are custom-designed to encase: (i) a central backbone made-up of a compression spring ($\Phi$ = 11 mm, h = 42 mm); (ii) rubber bands arranged in parallel at the outer extremity of the plates through dedicated grooves, and spaced radially at a distance of 60 degrees with-respect-to each other, and; (iii) tendon-guiding holes arranged at a spacing of 120 degrees. The combination of the compression spring and the rubber bands ensure that the system always remains in tension. One modular section, as shown in Fig.\ref{fig:robotic_platform}(A), is a serial concatenation of two activation units. It is actuated independently through a triad of fiber-based tendons arranged in parallel through the tendon-guiding holes. The overall system is made-up of three serially concatenated modular sections whose diameter decreases from base to tip, i.e., from 50 mm to 40 mm. In this way, the overall body is kinematically redundant, hollow, and lightweight. 
        
        \subsection{Actuation}
            The robot arm is mounted on a 3D printed rigid frame which encases nine motors (Dynamixel XM-430-W210-R, ROBOTIS Co. Ltd.), such that, the tendons are taut. In the initial configuration, the robot arm is oriented vertically upward, as illustrated in Fig.\ref{fig:robotic_platform}(C). By operating the motors at the position-level in the current arrangement, the system can generate two basic functionalities, i.e., omnidirectional bending and shortening, as discussed in next sections.
            
        \subsection{Sensing}
            Regarding the sensing capabilities, three proximity sensors (VL53L0X, STMicroelectronics) have been distributed around the circumference of the distal modular section, as shown in Fig.\ref{fig:robotic_platform}(D). Specifically, this sensor is a time-of-flight laser-ranging module which provides accurate absolute 1D distance measurement up to 2 m. It has a programmable I2C interface for device control and data transfer, which is sent by serial communication to the PC via a US cable

            Note that three retro-reflective markers have been radially distributed at the tip of each activation unit, thus, providing the pose of that activation unit in 3D cartesian space using vision-feedback. These sensors are employed for the sole purpose of monitoring the configuration of the robot arm in order to quantify further experimental results.      
            
    \section{Control Framework} \label{control_framework}
        The objective of this work is the development of a behaviour-based controller, which comprises of a bottom-up collection of behaviours. 

        \subsection{Primitive behaviours} \label{prim_behav}
            Primitive behaviours are behaviours that represent necessary functionalities, in the sense that each either achieves, or helps to achieve, a relevant goal that cannot be achieved without it by other members of that set. The controller comprises of the following set of primitive behaviours: 
            \begin{itemize}
                \item[(i)] \textit{bending in a principal direction} - it is well-known that a triad of radially arranged flexible actuators can generate six principal directions of bending, which can be achieved in one of two ways: (i) activating a single tendon in the proximal modular section, as well as, the corresponding tendons of the remaining distal modular sections that share a common actuation route, or; (ii) activating two adjacent tendons in the proximal modular section, as well as, the corresponding tendons of the remaining distal modular sections that share a common actuation route.
                \item[(ii)] \textit{resistance to bending in a principal direction} – for each configuration in (i), there will always be an antagonistic tendon configuration available whose activation will tend to bend the system in a direction opposite to (i). 
            \end{itemize}
            \begin{figure}[h!]
                \centering
                \includegraphics[width=\columnwidth]{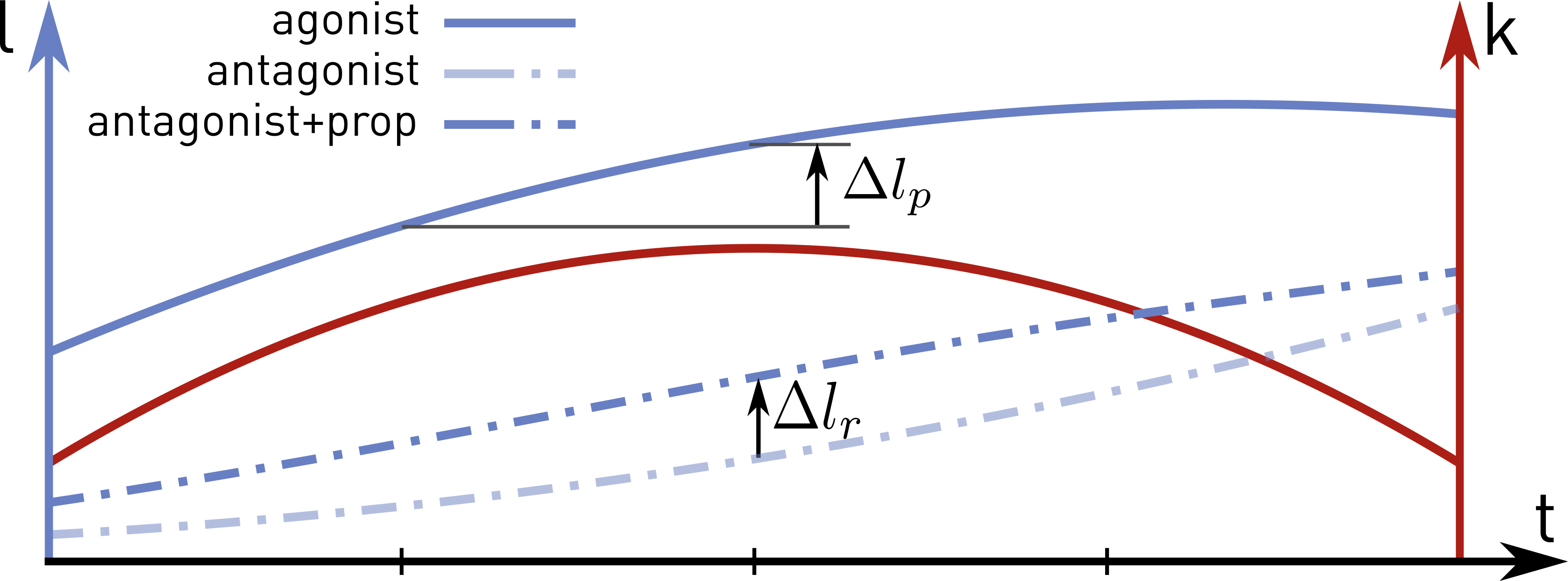}
                \caption{The antagonist restoring length is proportional to the agonist pulling length.}
                \label{fig:agonist_antagonist}
            \end{figure}
            Note that two modalities for antagonistic tendons actuation have been considered: pulled by a constant restoring length $\Delta l_r$, or proportionally to the agonists shortening $\Delta l_p$ according to the law $\Delta l_r = \alpha \Delta l_p$, where $\alpha \in [0, 1]$. If $\alpha = 0$, restoring will be null. If $\alpha = 1$, the arm will not bend but shorten. The restoring law changes proportionally to the curvature of the respective section: $\alpha = \kappa$, as shown in Fig.\ref{fig:agonist_antagonist}. Note that the need to incorporate this property has already been discussed in \cite{c9}. The curvature $\kappa$ of a section of diameter $\Phi$ is a function of the three tendon lengths $l_1, l_2, l_3$ \cite{c10}:
            \begin{equation}
                \kappa(l_1,l_2,l_3) = \frac{2 \sqrt{l_1^2 + l_2^2 + l_3^2 - l_1l_2 - l_1l_3 - l_2l_3}}{\Phi (l_1 + l_2 + l_3)}
            \end{equation}

        \subsection{Abstract behaviours} \label{abstract_behav}
            Abstract behaviours are behaviours which associate a more generalized set of activation conditions to trigger the primitive behaviours. Note that abstract behaviours allow for the modularization of behaviour-based networks, which is especially attractive to enable deliberative-like capabilities without foregoing the distributed nature of the controller \cite{c11}. The following abstract behaviours have been employed:
            \begin{itemize}
                \item[(i)] \textit{circular shift} - the sequential activation and deactivation of the tendons in the proximal modular section (as well as, the corresponding tendons of the remaining distal modular sections that share a common actuation route) along the cross-section of the robot arm. This behaviour generalizes two aspects of the primitive behaviour (i) from Section \ref{prim_behav}, the amplitude of the bend and the index of the actuator where the bend occurs. 
                \item[(ii)] \textit{enabling/disabling resistance to bending in principal direction} – this generalizes the activation/deactivation condition of primitive behaviour (ii) from Section \ref{prim_behav}, depending upon which kind of growth-driven behaviour is required, and discussed in the subsequent section. 
                \item[(iii)] \textit{learning from history} – as the system moves, its bending configuration, resistance to bending configuration, and sensory data is stored over time, forming a knowledge base over time. This generalizes two aspects of the primitive behaviour (i) from Section \ref{prim_behav}, the amplitude of the bend and the index of the actuator where the bend occurs. 
            \end{itemize}
            \begin{figure}[h!]
                \centering
                \includegraphics[width=\columnwidth]{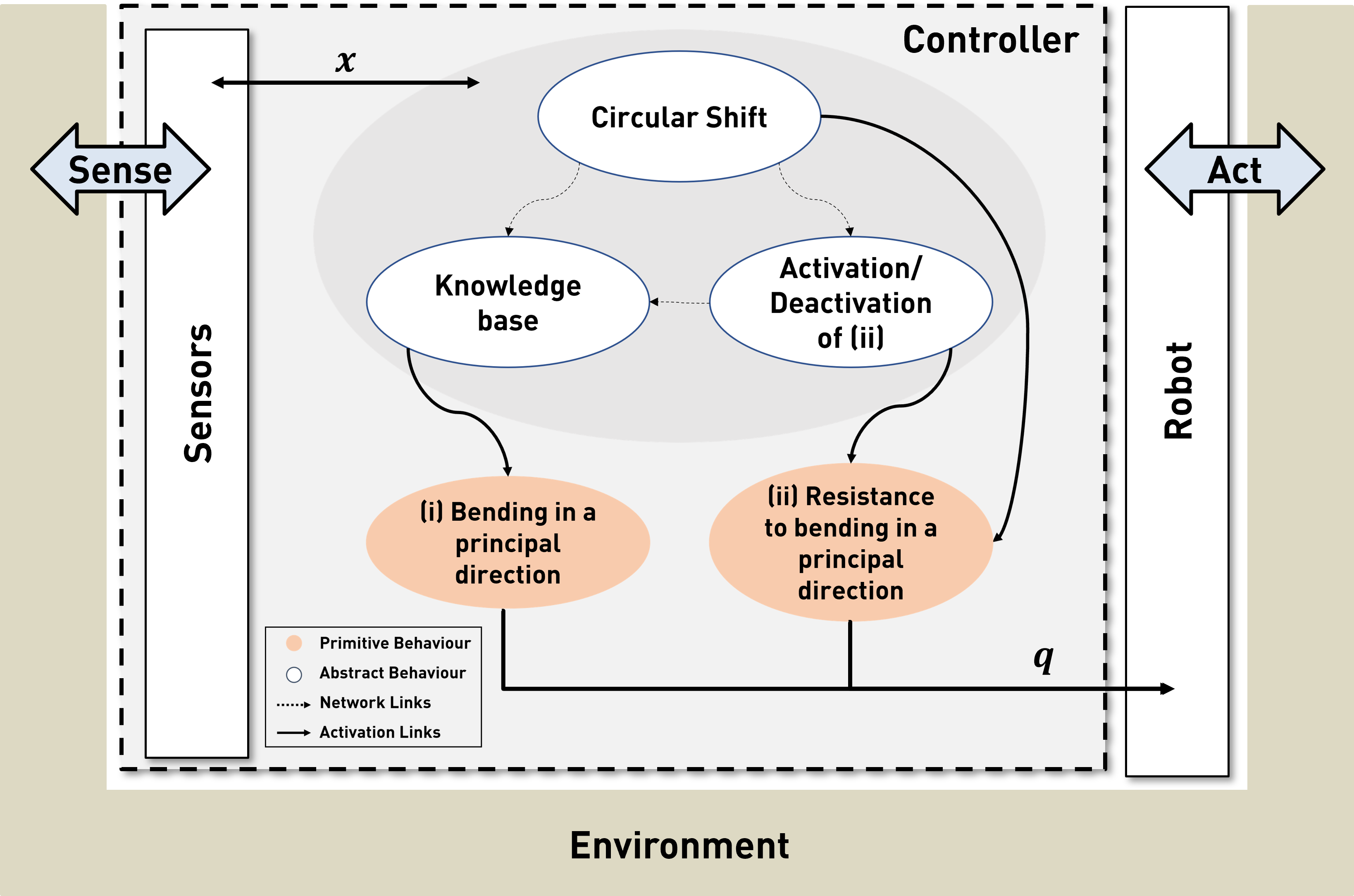}
                \caption{Overview of the behaviour-based control architecture.}
                \label{fig:control_architecture}
            \end{figure}
            
        \subsection{Control Architecture}
            The abstract behaviours and the primitive behaviours are combined in a bottom-up manner as shown in Fig. \ref{fig:control_architecture}. The controller generates the growth-driven-like movements in the continuum arm, as follows:
            \begin{itemize}
                \item[(i)] Exploration-phase: a circular shift (refer to (i) of Section \ref{abstract_behav}) of the principal direction of bending (refer to (i) in Section \ref{prim_behav}) is applied while the resistance to the principal direction of bending is disabled (refer to (ii) in Section 3.2). As is shown in the results, this combination of behaviours should generate a circumnutation-like rotation. The traversed configurations and respective distance to the target are saved in a knowledge base. The total duration of this functionality will be defined by the number of steps in one complete rotation and the discretization of the amplitude. Note that the resistance value can be increased, and will be investigated in the future works. 
                \item[(ii)] Reaching-phase: after the exploratory phase has competed, the controller searches through its knowledge base (refer to (iii) from Section \ref{abstract_behav}) to find the configuration that has the least distance with the external stimulus. Then it applies the corresponding bending configuration (refer to (i) in Section \ref{prim_behav}). In this phase, we also investigate the effect on the curvature by activating the resistance to the bending (refer to (ii) in Section \ref{abstract_behav} and (i) in Section \ref{prim_behav}).
            \end{itemize}
        
    \section{Results} \label{results}
        \subsection{Continuum arm characterization}
            \begin{figure}[h!]
                \centering
                \includegraphics[width=.9\columnwidth]{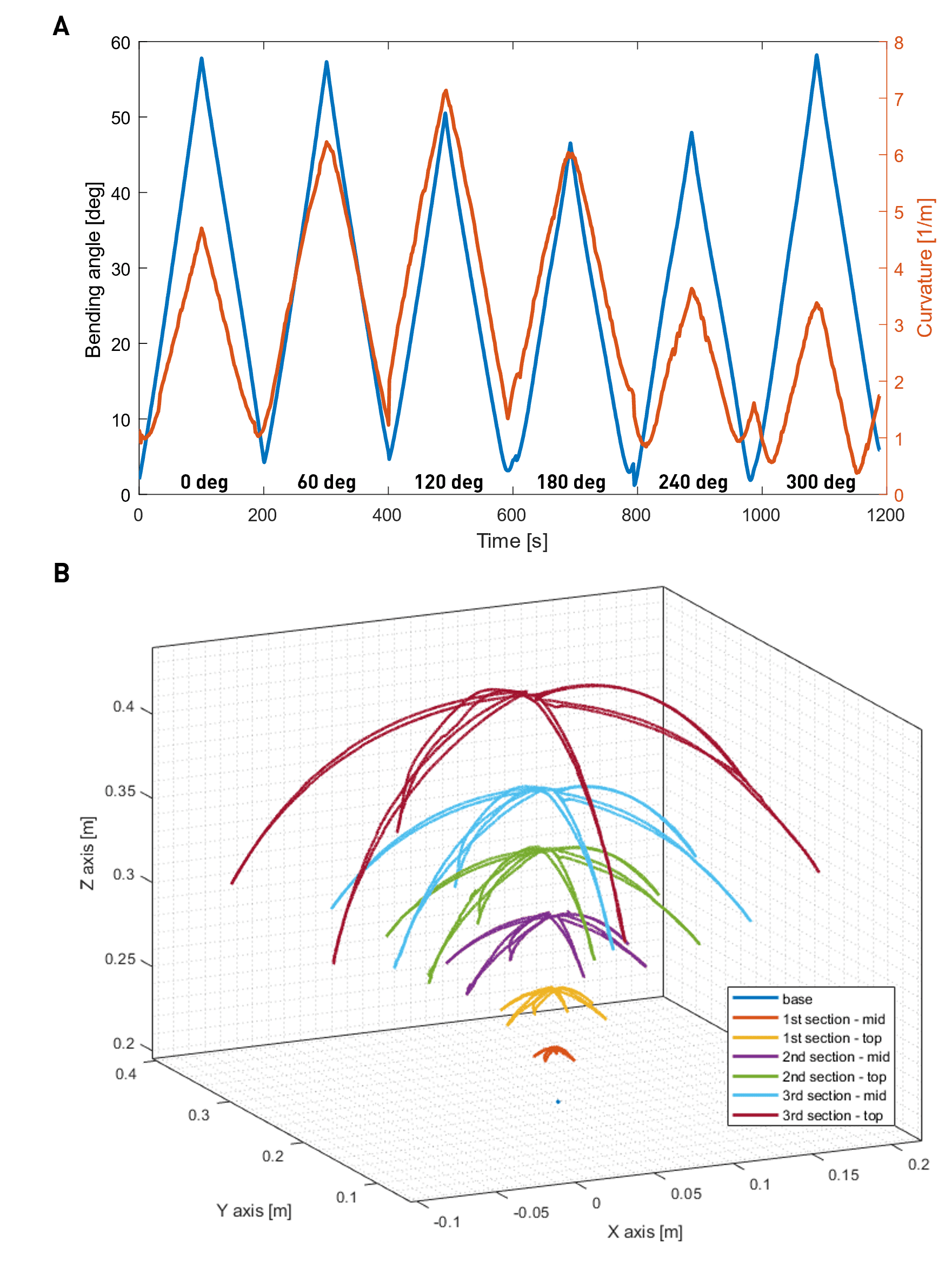}
                \caption{The principal bending directions are identified by pulling respective tendons. (A) Comparison between bending angle and structure curvature over all the bending directions. (B) Segment trajectories while moving along the bending directions.}
                \label{fig:characterization}
            \end{figure}
            In order to visualize the behaviour, we employ retroreflective markers distributed across the body at the tip of each module, i.e., at six discretised locations. They are able to measure the 3D position in Cartesian space through vision feedback system (BTS, Inc.), thereby enabling to capture the progressive motion of the system during bending. In particular, the motion of each module is represented in a different color, as illustrated in Fig.\ref{fig:characterization}(B). In general, the principal directions of bending are spaced 60 degrees apart from each other. For each principal direction of bending, it can be seen that the bending pattern remains consistent for each module. Furthermore, from the initial configuration of the robot along the positive z-axis the robot rotates through a bending angle of approximately 60 degrees, as shown in Fig.\ref{fig:characterization}(A). Also, the arm shows a low hysteresis while moving back and forth along the same direction. Interestingly, the change in the bending angle from the initial configuration towards the final configuration exhibits a linear trend. The change in bending angles for each direction varies due to fabrication inconsistencies and gravitational effects.
            
        \subsection{Experimental target positioning}
\begin{figure}[h!]
                \centering
                \includegraphics[width=.8\columnwidth]{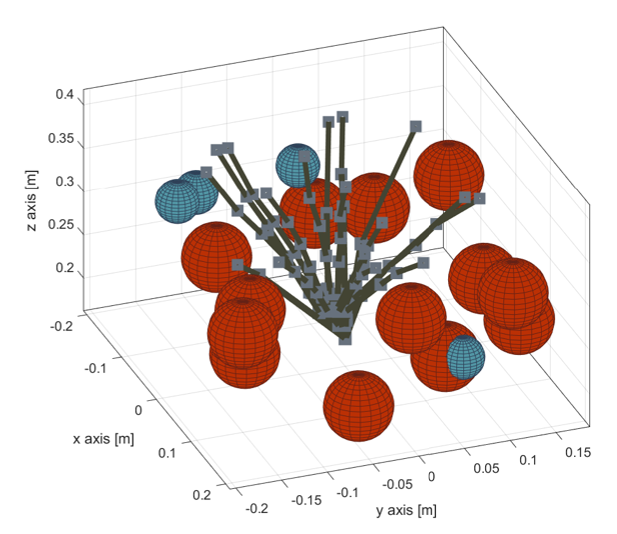}
                \caption{Targets are placed around the continuum arm to show the controller invariancy to their position. The closest reaching continuum arm configuration is reported for each target. For sake of visibility, targets are represented with half of their actual diameter.}
                \label{fig:targets}
            \end{figure}

            Two spherical targets with different diameters (i.e., $\Phi_1$ = 160 mm, $\Phi_2$ = 100 mm) are used to show how the controller behaves in different situations. Moreover, targets positioning is chosen to cover the whole robot workspace, to demonstrate that the strategy is invariant to their position with respect to the arm. The two targets are positioned in 5 and 13 different locations respectively, covering different heights, angles and distances, as shown in Fig.\ref{fig:targets}.

        \subsection{Exploration phase}
\begin{figure}[h!]
                \centering
                \includegraphics[width=1\columnwidth]{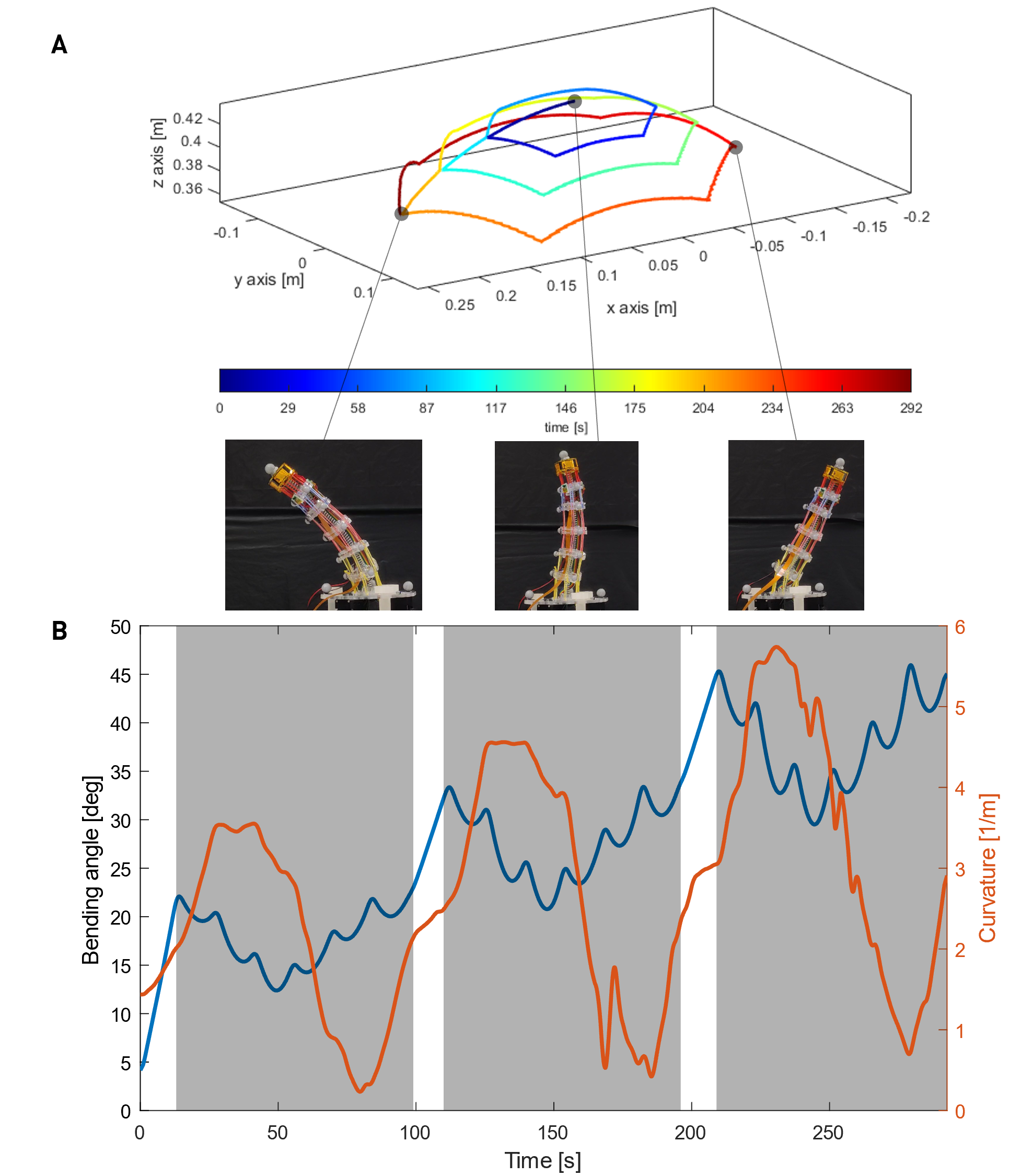}
                \caption{The exploration phase is inspired to plants’ circumnutation. (A) Tip trajectory over time during exploration. (B) Bending angle and curvature of continuum arm structure during exploration.}
                \label{fig:exploration}
            \end{figure}

            The exploration phase is inspired by plants’ circumnutation. In order to characterize a circumnutation-like behaviour, we again employ retro-reflective markers, however, this time they are only placed at the end-effector of the robot. A circular shift of bending is applied from actuators in a counter clockwise direction, with an increasing amplitude. Speficially, the amplitude of the bending is gradually increased after one complete rotation in 100 steps. Fig.\ref{fig:exploration} depicts the resulting trajectory of the system, which is a closed loop orbit in phase space, but with slight inconsistencies in the curvature.       

            \subsubsection{Proximity observation space}
                \begin{figure}[h!]
                    \centering
                    \includegraphics[width=\columnwidth]{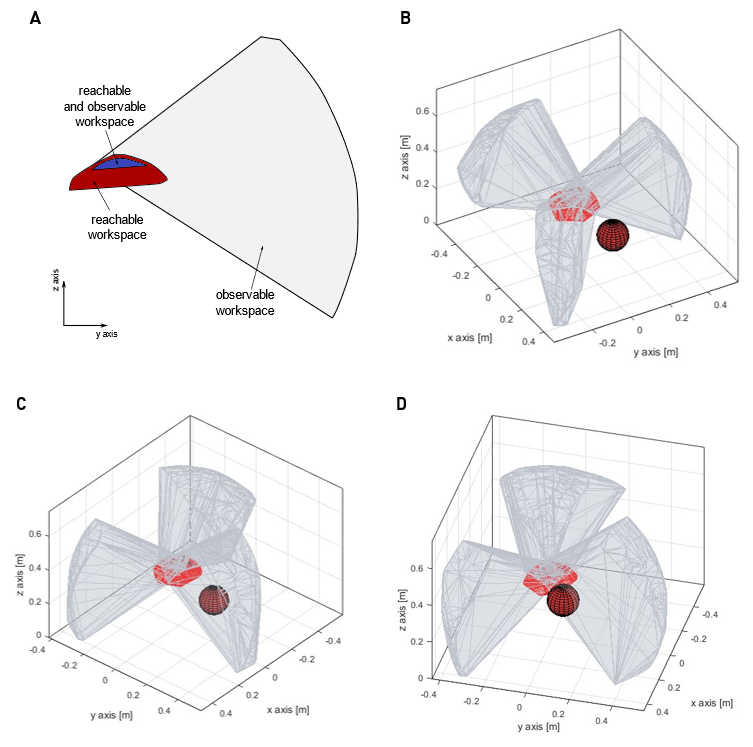}
                    \caption{Proximity sensors detect the target positioning in the robot’s surroundings. (A) For sake of simplicity, the plane of a single principal bending direction is considered. The reachable and observable workspaces are identified. (B) The target is out of the proximity observation space, so it is not detected. (C) The target is observed by proximity sensors. (D) The target can be both observed and reached by the continuum arm.}
                    \label{fig:observation}
                \end{figure}
                
                Proximity sensors can detect a target in the continuum arm’s surroundings. The robot’s workspace is limited with respect to proximity observable space, so it can be helpful to identify some sub-spaces, as shown in Fig.\ref{fig:observation}(A). We can distinguish the proximity observation space and the workspace that is physically reachable by the continuum arm; moreover, their intersection generates the workspace in which the target can be both observed and reached.
                
                In particular, three different situations can be experiences with such robotic system: Fig.\ref{fig:observation}(B), the target is out of the observable workspace by proximity sensor, so it will never be detected and approached by the continuum arm; Fig.\ref{fig:observation}(C), the target is partially or entirely within the observable workspace, so the continuum arm can detect and move towards it; Fig.\ref{fig:observation}(D), the target is also intersecting the robot’s reachable workspace, so the arm can finally touch it.
                
\begin{figure}[h!]
                    \centering
                    \includegraphics[width=.9\columnwidth]{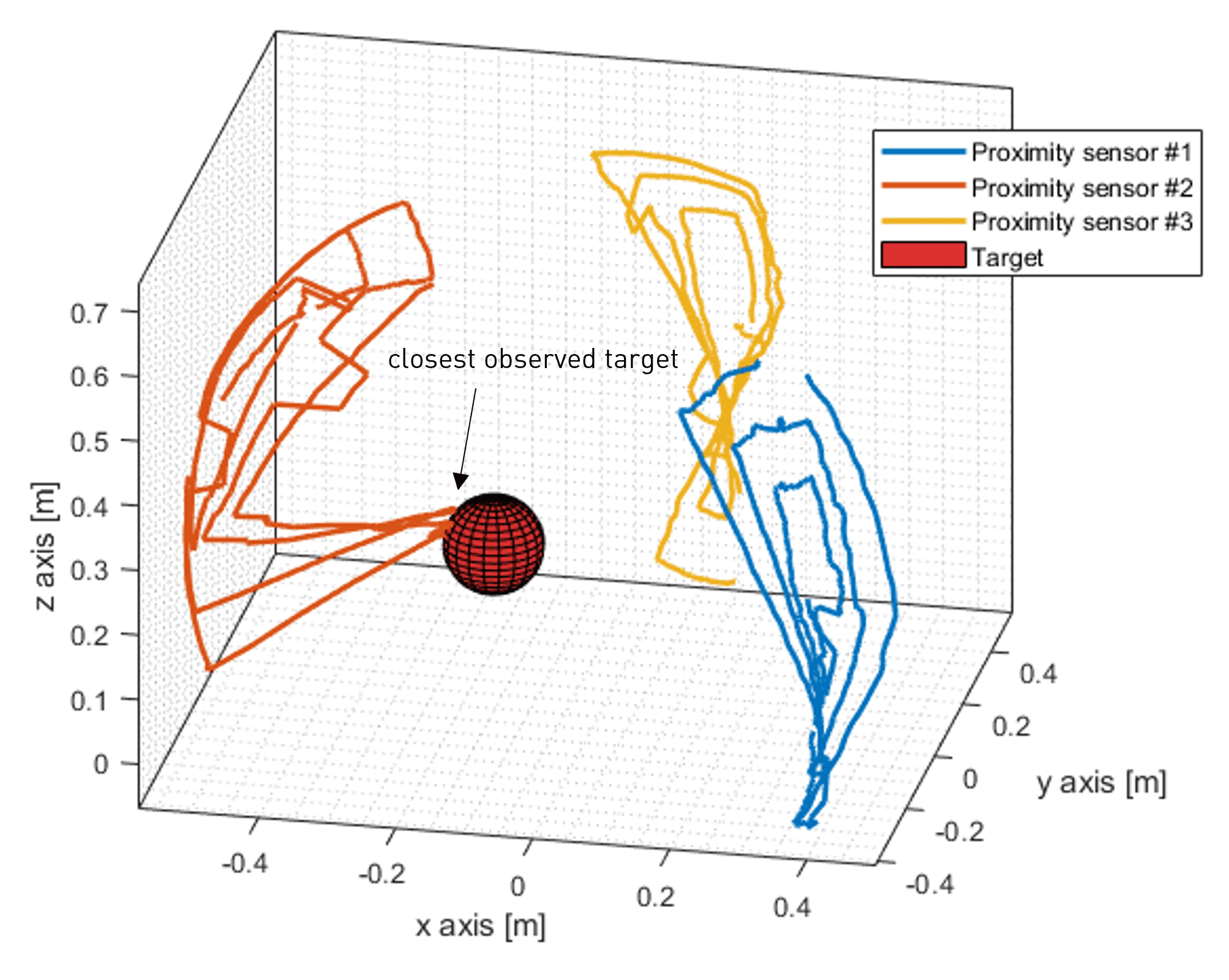}
                    \caption{The proximity sensor output is compared with respect to the actual target positioning.}
                    \label{fig:closest_observation}
                \end{figure}

                After the exploration phase, the robot moves towards the configuration closest to the target. In Fig.\ref{fig:closest_observation}, the proximity sensors output is mapped to the 3D space and compared with the actual target position. Proximity sensors correctly identify the closest target position within the observation space.

        \subsection{Reaching phase}
            If the target is identified, the robot approaches towards it for reaching. We tested different tendon actuations to find the best reaching performance, recurring to antagonist tendons pulling. In particular, we analysed cases in which they are shortened by a constant factor or proportional to the continuum arm’s curvature. 

            \footnotesize
            \begin{table}[h!]
                \begin{tabular}{|m{.17\columnwidth}|m{.2\columnwidth}|m{.21\columnwidth}|m{.21\columnwidth}|}
                \hline
                \textbf{Target}          & \textbf{No VS constant pulling} & \textbf{No VS proportional pulling} & \textbf{Constant VS proportional pulling} \\ \hline
                Not   reachable & -0.11 $\pm$ 1.37 mm     & -0.67 $\pm$ 0.81 mm         & -0.56 $\pm$ 0.58 mm               \\ \hline
                Reachable       & 2.77 $\pm$ 0.57 mm      & -0.53 $\pm$ 0.49 mm         & -3.31 $\pm$ 0.32 mm               \\ \hline
                \end{tabular}
                \caption{Comparison among the three reaching strategies. For both reachable and not reachable targets, no antagonist pulling, constant and proportional pulling are considered. For each trial, the best reaching performance for all the cases is identified. The comparison quantifies how much one strategy outperforms the other.}
                \label{tab:reaching}
            \end{table}
            \normalsize

            The table Tab.\ref{tab:reaching} shows the comparison between no, constant and proportional antagonist actuation, based on the distance between the continuum arm and the target. The trials have been clustered with respect to reachable and not reachable targets. For both situations, the proportional antagonist pulling outperforms the other strategies. For not reachable targets, all the strategies allow to get the same best optimal configuration, but the proportional case allows for lower variations around it. For reachable targets instead, proportional pulling permits to get an overall better configuration than other cases; deviations from the mean are lower than the not reachable case, due to higher proximity sensors reliability.

\begin{figure}[h!]
                \centering
                \includegraphics[width=\columnwidth]{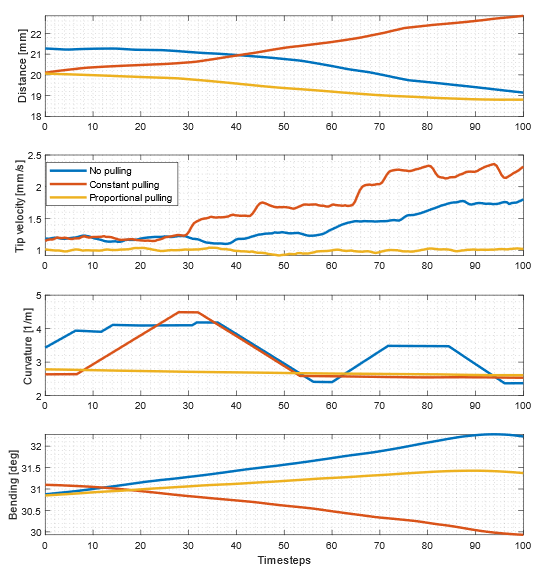}
                \caption{The three reaching modalities (i.e., no antagonist pulling, constant pulling, proportional pulling to curvature) of a single reaching trial are compared. The distance to the target, the tip velocity, and the continuum arm’s curvature and bending angle are considered as useful observations to estimate the performance of each modality and to enable their comparison.}
                \label{fig:reaching_trial}
            \end{figure}

            For sake of simplicity, the evolution of a single reaching trial is reported in Fig.\ref{fig:reaching_trial}. As anticipated by previous results, the proportional pulling has the best performance in terms of distance from the target. Moreover, it allows for constant tip velocity, continuum arm’s curvature and bending angle. It means that the proportional pulling generates a much more stable control of the arm than other cases, while ensuring the best reaching performance.

    \section{Discussion} \label{discussion}
        Results can be summarized concerning the characteristics of the robotic arm and its movement capabilities, the execution of programmed routines and the ability to perceive its surroundings via proximity sensors.

        The motion of the compliant continuum arm is characterised by its principal bending directions, generated by the activation of one or two tendons simultaneously to bend in the respective direction. The robot's bending angle and curvature characteristics vary across different directions due to manufacturing reasons, but this can be neglected during the analysis of the reaching task. In particular, the arm can describe a bending angle of approximately 60 degrees from the normal in a uniform manner, but the same does not apply to the curvature, which oscillates towards zero depending on the chosen direction.

        The non-isotropically behaviour of the robotic platform and the demonstration of the controller's independence of the relative position between robot and target is addressed by employing the choice of target positions in space to cover the entire workspace, also making use of spherical targets of varying diameters.

        Taking inspiration from plants and their growth mechanism, we implemented two behaviours that recall the plants' circumnutation for exploration purposes and the tropism to enable reaching. 

        The exploration aims to locate the position of the target in space using solely the information from proximity sensors. Each exploration cycle is characterised by a constant bending angle which is proportional to its diameter. Depending on the position and size of the target, it can be located in a different exploration phase. This characteristics is closely related to the proximity observation cone of the robot, so we can identify three areas of the workspace have been identified: (i) unobservable area, in which the target will not be observed by the robot at all; (ii) observable but unreachable area, in which the robot can locate the target but can not reach it employing the tip of the arm; (iii) observable and reachable area, in which the robot and the target manage to be in contact with each other. In addition, the proximity observation cone presents many blind views which could be diminished by increasing the number of employed sensors, or by exploiting a different arrangement of the tendons.

        The controller has always been consistent in exploration results, identifying the configurations of the arm closest to the target if observable, otherwise obtaining the wrong configurations if the target was not observable. Furthermore, measurements provided by proximity sensors are consistent with what is observed through the reconstruction of the robot backbone using a marker-based motion capture system. Being an online control technique, it fares better than existing solutions to external disturbances, which is essential for the target application.

        Finally, after identifying the correct configuration, the arm is free to reach the point estimated to be closest to the target. Several reaching implementations have been proposed, which rely on the concurrent actuation of agonist and antagonist tendons to generate motion but with the lowest curvature of the robot's shape. The case in which the opposing actuators are not actuated is used as a ground for comparison. The pulling of the antagonist actuators based on proprioception, as it happens by analogy in plants, generated the best performance in terms of target reachability in position and balance around it.

    \section{Conclusion and future works} \label{conclusion}
        For a continuum arm, a plant-inspired reaching behaviour is implemented through embedded sensing and a distributed control architecture. Task control is implemented without any internal model representation of the robot, instead relying solely on sensory data. During the exploration phase, proximity sensing allows the robot to find the target location, and proprioception allows the robot to achieve better reaching performance than in other cases. A similar approach can be implemented on a soft arm with common structural features to the continuum arm employed in this work.

        This implementation is a step towards the use of embedded sensing and distributed control strategies in continuum and soft robots. It enables robotic arms to perform reaching tasks out of the laboratory environment, dealing with dynamic changes. Further improvements of this work might consider to increase the cardinality of proximity sensors to cover the whole observable area, and to rely on contact sensing while moving close to the target surface. Different tendons arrangements can be examined to obtain different behaviours that will improve task performance. Furthermore, the primitive reaching mechanism could be used to test target tracking strategies or to see if it is robust to external disturbances that change the internal structure of the robot. 
    


\begin{thebibliography}{99}
        \bibitem{c1} C. Laschi, B. Mazzolai, and M. Cianchetti, 2016. Soft robotics: Technologies and systems pushing the boundaries of robot abilities, \textit{Science Robotics}, 1(1)
        \bibitem{c2} I.D. Walker, 2013. Continuous backbone “continuum” robot manipulators. \textit{Isrn robotics}, 2013
        \bibitem{c3} A. Trewavas, 2005. Plant intelligence, \textit{Naturwissenschaften}, 92(9), pp.401-413
        \bibitem{c4} A. Trewavas, 2003. Aspects of plant intelligence, \textit{Annals of botany}, 92(1), pp.1-20. 
        \bibitem{c5} R.R. Murphy, 2019. Introduction to AI robotics, \textit{MIT press}
        \bibitem{c6} R.A. Brooks, and J.H. Connell, 1987. Asynchronous distributed control system for a mobile robot, in \textit{Mobile Robots I }(Vol. 727, pp. 77-84). International Society for Optics and Photonics 
        \bibitem{c7} R.C. Arkin, 1998. Behavior-based robotics. \textit{MIT press}
        \bibitem{c8} M. Rivière, J. Derr, and S. Douady, 2017. Motions of leaves and stems, from growth to potential use. \textit{Physical biology}, 14(5), p.051001
        \bibitem{c9} E. Donato, Y.T. Ansari, C. Laschi, and E. Falotico, 2022. To Enabling Plant-like Movement Capabilities in Continuum Arms, \textit{Proceedings of 2022 I-RIM Conference}, 6th-8th October 2022, Rome, Italy
        \bibitem{c10} R.J. Webster III, and B.A. Jones, 2010. Design and Kinematic Modeling of Constant Curvature Continuum Robots: A Review, \textit{The International Journal of Robotics Research}, 29(13), pp. 1661-1683
        \bibitem{c11} M.J. Mataric, and F. Michaud, 2008. Behavior-Based Systems, \textit{Springer Handbook of Robotics}, pp 891–909
    \end{thebibliography}
\end{document}